\documentclass[sigconf]{acmart}

\setcopyright{rightsretained}
\acmConference[SIGSPATIAL '26]{The 34th ACM SIGSPATIAL International Conference on Advances in Geographic Information Systems}{November 3--6, 2026}{Riverside, CA, USA}
\acmYear{2026}
\copyrightyear{2026}
\acmISBN{}
\acmDOI{}
\renewcommand{\footnotetextcopyrightpermission}[1]{}

\usepackage{booktabs}
\usepackage{graphicx}
\usepackage{multirow}
\usepackage{tikz}
\usetikzlibrary{arrows.meta,positioning,fit,backgrounds}
\tikzset{
  stage/.style={draw, rounded corners, align=center, font=\scriptsize,
                text width=1.85cm, minimum height=0.95cm, fill=black!4, inner sep=2pt},
  io/.style={draw, align=center, font=\scriptsize, text width=1.85cm,
             minimum height=0.95cm, fill=black!10, inner sep=2pt},
  flow/.style={-{Stealth[length=2mm]}, thick},
}

\begin{document}

\title{MapReason-OSM: Can Vision-Language Models Make Graph-Verifiable Mobility Decisions from Street Maps?}

\author{Srinivas Venkatanarayanan\textsuperscript{1,2} \qquad Clement Pakkam Isaac\textsuperscript{1,3}}
\affiliation{%
  \institution{\textsuperscript{1}NVIDIA \quad \textsuperscript{2}University of Central Florida \quad \textsuperscript{3}University of South Florida}
  \country{USA}}

\begin{abstract}
Vision-language models (VLMs) are increasingly used to read maps for logistics, delivery, and accessible navigation, where the output is an actionable decision (a route, a pin, a parking choice) that must respect the road network. Yet most map benchmarks grade free-text or multiple-choice answers that cannot be verified against the underlying graph. We present \textbf{MapReason-OSM}, a benchmark and evaluation harness for graph-verifiable mobility decisions on self-rendered OpenStreetMap panels. We render fixed-style maps for ten U.S. downtowns at two aligned zoom scales, overlay a consistent marker grammar, and pair each panel with a hidden street graph and exact oracles, yielding 6{,}000 instances (12{,}000 panels across the two zooms) over 12 routing, facility-location, and visual-disambiguation tasks. Models return structured decisions that we snap back to the graph and score for validity, legality, optimality, and constraint satisfaction, plus \emph{cross-zoom consistency}. Across seven VLMs, models read maps and route simply but fail at graph-cost reasoning (single-facility pin placement is near chance even for frontier reasoning models), and are frequently scale-inconsistent. We release the benchmark, harness, and deterministic generator.
\end{abstract}



\maketitle

\begingroup\small\itshape\noindent
Preprint. This work has been submitted to the ACM SIGSPATIAL 2026 Industrial Track and is currently under review; this version may differ from the final published paper. Code, data, and the deterministic generator: \url{https://github.com/Vi-Sri/mapreason-osm}.
\par\endgroup
\medskip

\section{Introduction}
Vision-language models (VLMs)~\cite{openai2023gpt4,geminiteam2023gemini,liu2023llava,bai2023qwenvl} are being deployed on maps for logistics, last-mile delivery, field operations, and accessible navigation. In each of these settings the output is not free text but an \emph{actionable decision}: a route a vehicle will drive, a pin a driver will navigate to, a parking lot a delivery will use. Such decisions are only useful if they are \emph{legal} (they follow real, connected streets and obey one-way, closure, and accessibility constraints) and \emph{near-optimal} (short under network distance). Yet most map-and-VLM benchmarks grade free-text or multiple-choice answers and therefore cannot verify any of these properties against the actual road network.

This gap matters precisely where the industry wants to deploy these models. A model that ``reads'' a map well (identifying labels, icons, and spatial relations) may still propose a route that traverses non-adjacent streets, drives the wrong way down a one-way, or selects a facility that is far by road despite looking close on the image. Distinguishing genuine spatial \emph{reasoning} from competent map \emph{reading} requires scoring against a ground-truth graph, not against text similarity.

We introduce \textbf{MapReason-OSM}, a benchmark and evaluation harness that makes this distinction measurable.\footnote{Code, data, and the deterministic generator are available at \url{https://github.com/Vi-Sri/mapreason-osm}.} Every instance is a street-map panel that we render ourselves from OpenStreetMap (OSM) vector data (never scraped tiles), so that we control the cartographic style, the visible-marker grammar, and the aligned multi-zoom extents. Behind each panel sits a hidden street graph and an exact oracle. The model, seeing only the image and a prompt, returns a structured JSON decision; we snap its marker references back to the hidden graph, induce the decision on the graph, and score it for schema validity, legality, optimality, and constraint satisfaction.

Our contributions are:
\begin{itemize}
\item \textbf{A graph-grounded VLM benchmark on self-rendered OSM maps} with an exact symbolic scoring chain (schema validity, then legality, optimality, and constraint satisfaction): 6{,}000 instances and 12{,}000 panels across ten U.S. downtowns and 12 tasks, fully regenerable from a fixed seed.
\item \textbf{Three evaluation axes not previously combined}: cross-zoom consistency, accessibility-aware (step-free) routing, and graph-grounded facility-location pin placement, alongside a symbol-grounded industrial task family (drawn closures, hazard zones, loading docks, attributed parking).
\item \textbf{A reproducible, massively parallel harness} and a seven-model study (five fast, two frontier) showing a sharp perception-vs-reasoning gap: VLMs read maps and route simply, but graph-cost reasoning (most cleanly, single-facility placement) stays near chance even for frontier reasoning models, and decisions are frequently scale-inconsistent.
\end{itemize}

\textbf{Industrial relevance.} The tasks are deliberately drawn from real logistics and assistive-mobility operations (pickup-pin selection, depot siting, loading-dock and parking choice, hazard- and closure-aware rerouting, and step-free routing), so the benchmark measures exactly the decisions a deployed map-reading assistant must get right. Our headline result is a transferable lesson for practitioners: today's VLMs can be trusted to read a labeled map but not to optimize over its road network, and their answers can change with zoom; both are concrete, measurable failure modes to guard against in production.

The paper is structured as follows. Section~\ref{sec:related} positions MapReason-OSM against prior map and geospatial benchmarks. Section~\ref{sec:bench} describes the data substrate (self-rendered panels, marker grammar, hidden graph and oracles, deterministic generation). Section~\ref{sec:tasks} defines the 12 tasks and Section~\ref{sec:protocol} the scoring protocol. Sections~\ref{sec:setup} and~\ref{sec:results} present the experimental setup and results, and Sections~\ref{sec:discussion}--\ref{sec:conclusion} discuss open challenges, limitations, and conclusions.

\section{Related Work}
\label{sec:related}
\textbf{Map reading and reasoning for (V)LLMs.} A growing line of benchmarks asks whether multimodal models can read maps. MapBench evaluates pixel-space path finding on illustrated, human-readable tourist maps via a map-space scene graph~\cite{xing2025mapbench}; ReasonMap targets fine-grained visual reasoning over high-resolution transit maps~\cite{feng2026reasonmap}; MapTab probes multi-criteria route planning over heterogeneous graphs by fusing map images with attribute tables~\cite{shang2026maptab}; and MapEval spans textual, API, and visual map question answering, finding that no foundation model exceeds 67\% and all trail humans by over 20 points~\cite{dihan2025mapeval}. The community's interest in this direction is evident at SIGSPATIAL itself, which recently introduced cartographic and open-domain map question-answering benchmarks for (V)LLMs~\cite{ung2025cartomapqa,li2025mapqa}. These benchmarks predominantly grade free-text or multiple-choice answers, or operate on illustrated/transit maps. MapReason-OSM differs by (i) using \emph{self-rendered OSM street} panels, (ii) scoring \emph{structured} decisions against a \emph{hidden} street graph and exact oracle, and (iii) measuring cross-zoom consistency, accessibility-aware routing, and graph-grounded facility-location placement in a single, exactly-scored suite. We deliberately avoid first-of-its-kind claims; our contribution is the combination of axes and the industrial task framing.

\textbf{Multimodal reasoning, visual perception, and graph reasoning.} Beyond maps, a broad suite of benchmarks probes whether multimodal models can reason over structured visual content: expert, multi-discipline understanding~\cite{yue2024mmmu}, mathematical reasoning in visual contexts~\cite{lu2024mathvista}, and question answering over charts and documents~\cite{masry2022chartqa,mathew2021docvqa}. A recurring finding is that VLMs ``can see but not perceive''~\cite{fu2024blink}: they handle textual cues yet falter on core perception, spatial relations, and object orientation~\cite{liu2023vsr,chen2024spatialvlm}, and they hallucinate ungrounded content~\cite{li2023pope}. In parallel, work on graph reasoning shows that language models struggle to solve graph problems posed in natural language and are sensitive to how the graph is encoded~\cite{wang2023nlgraph,fatemi2024talklikegraph}. MapReason-OSM sits at the intersection: it demands both fine-grained visual perception of a cartographic scene \emph{and} cost-aware reasoning over the graph that the scene depicts, and it scores the latter exactly rather than as free text.

\textbf{Geospatial knowledge and reasoning in LLMs.} LLMs encode substantial geographic knowledge, which can be extracted with auxiliary OSM context~\cite{manvi2024geollm} but is geographically biased~\cite{manvi2024geobias} and uneven across factual and interpretive geographic tasks~\cite{roberts2023gpt4geo}. Multi-task studies show text-only LLMs are weak at spatial tasks such as route planning, with large prompt-strategy effects~\cite{xu2024spatialtasks}. We complement these text-centric analyses by isolating the \emph{visual} graph-grounded setting.

\textbf{Route-planning agents and benchmarks.} MobilityBench evaluates tool-mediated LLM route-planning agents in a deterministic API-replay sandbox over real Amap queries~\cite{song2026mobilitybench}; GeoBenchX benchmarks LLM tool-calling on multi-step geospatial tasks~\cite{krechetova2025geobenchx}; and agentic LLM frameworks turn heterogeneous urban data into planner-facing spatial intelligence~\cite{xu2025agentic}. These are \emph{tool-augmented} or \emph{text/data} pipelines; our Pure Visual Decision track requires a graph-checkable decision read directly from a rendered map, without tools or tabular access.

\textbf{Mobility, accessibility, and data substrate.} LLMs have been used to synthesize travel surveys for urban mobility assessment~\cite{bhandari2024urban}, and accessibility-first routing systems such as AccessMap demonstrate the importance of step-free, profile-aware paths~\cite{bolten2019accessmap}; we make accessibility-aware routing a benchmarked VLM task. Large reproducible spatio-temporal resources~\cite{yin2025xxltraffic} reflect the community's emphasis on rigorous benchmarks. We build street graphs and features with OSMnx~\cite{boeing2017osmnx} from OpenStreetMap~\cite{openstreetmap}, rendering our own panels (rather than scraping tiles) in compliance with the OSM tile usage policy.

\section{The MapReason-OSM Benchmark}
\label{sec:bench}
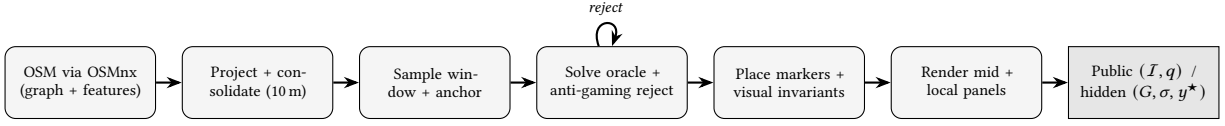
\begin{figure*}[t]
\centering
\begin{tikzpicture}[node distance=0.42cm and 0.34cm]
  \node[stage] (g1) {OSM via OSMnx (graph + features)};
  \node[stage, right=of g1] (g2) {Project + consolidate (10\,m)};
  \node[stage, right=of g2] (g3) {Sample window + anchor};
  \node[stage, right=of g3] (g4) {Solve oracle + anti-gaming reject};
  \node[stage, right=of g4] (g5) {Place markers + visual invariants};
  \node[stage, right=of g5] (g6) {Render mid + local panels};
  \node[io, right=of g6] (g7) {Public $(\mathcal{I},q)$ \,/\, hidden $(G,\sigma,y^\star)$};
  \draw[flow] (g1)--(g2); \draw[flow] (g2)--(g3); \draw[flow] (g3)--(g4);
  \draw[flow] (g4)--(g5); \draw[flow] (g5)--(g6); \draw[flow] (g6)--(g7);
  \draw[flow] (g4) to[out=110,in=70,looseness=6] node[above,font=\scriptsize\itshape]{reject} (g4.north);
\end{tikzpicture}
\caption{Dataset-generation workflow: deterministic, anti-gaming construction from OSM to paired public panels and the hidden graph/oracle. The \emph{reject} self-loop discards instances whose visual shortcut already matches the oracle (Section~\ref{sec:bench}).}
\label{fig:gen}
\end{figure*}

At a high level, an instance is a tuple $(\mathcal{I}, q, G, \sigma, y^\star)$: one or more rendered panels $\mathcal{I}$, a natural-language prompt $q$, a hidden street graph $G$, a snapping function $\sigma$ that maps each visible marker ID to a graph node, and an oracle answer $y^\star$. Only $(\mathcal{I}, q)$ is shown to the model; $(G, \sigma, y^\star)$ are held out for scoring. Figure~\ref{fig:gen} shows the deterministic dataset-generation workflow; the matching graph-grounded evaluation workflow is given later in Figure~\ref{fig:eval} (Section~\ref{sec:protocol}). We describe each component below.

\subsection{Self-rendered panels}
Each panel is a $1024\times1024$ north-up PNG. We acquire the street network and point/area features for each metro with OSMnx~\cite{boeing2017osmnx}, project to a metric UTM CRS, and render over a fixed CartoDB Voyager raster basemap via \texttt{contextily}; a faint, translucent single-color tracing of the routable graph edges (alpha $0.2$) is drawn beneath the markers so junctions and segments read clearly and markers visibly connect to the network. Rendering our own panels (rather than scraping raster tiles) gives us control over the cartographic style, the marker grammar, and the aligned multi-zoom extents, and keeps us compliant with the OSM tile usage policy. Every panel stores an affine transform between pixel and UTM coordinates and carries an attribution stamp (``\textcopyright{} OpenStreetMap contributors, \textcopyright{} CARTO''). Each instance is rendered at two \emph{aligned} zoom scales sharing one center: a \emph{mid} panel (half-extent $500$\,m; $\approx$1\,km across, for street selection and POI context) and a \emph{local} panel (half-extent $175$\,m; $\approx$350\,m across, for turns, curbs, and fine pin placement). Because the two panels share a center and the local extent is nested inside the mid extent, and because every task-critical marker is placed within a $130$\,m-radius window of that center (well inside both extents), the same markers and cues appear on \emph{both} panels. This guaranteed overlap is what makes cross-zoom consistency meaningful: the model is shown identical evidence at two scales, so a change in its answer reflects scale-sensitivity rather than missing information.

\subsection{Street graphs: drive vs.\ walk}
The hidden graph is task-appropriate. Vehicle and logistics tasks (legal/one-way/closure/symbol-constrained/POI routing, pin placement, service-area siting, reachability, attribute-aware parking, loading-dock selection) use the \emph{directed drive} network, so that one-way restrictions and turn legality are meaningful. Pedestrian tasks (accessibility/step-free routing and visual disambiguation) use the \emph{walk} network, which is denser and includes footways and stairs. One-way compliance specifically requires the directed drive graph: legality is defined by edge direction. Graphs are projected and intersection-consolidated at a $10$\,m tolerance to remove dual-carriageway artifacts; node attributes carry coordinates and, where available, OSM tags (\texttt{oneway}, \texttt{highway}, \texttt{wheelchair}, \texttt{kerb}, \texttt{incline}, \texttt{smoothness}). We additionally compute per-window morphology statistics (orientation entropy, intersection density, circuity) so instances can be stratified across grid, organic, and sparse street fabrics.

\subsection{Visible-marker grammar and symbol overlays}
A fixed grammar makes the on-panel labels unambiguous and machine-checkable. Core markers are colored dots with a bold printed ID: \texttt{A} (green) = start, \texttt{E} (red) = goal, \texttt{W} (slate) = required waypoint, \texttt{I1, I2,\dots} (purple) = junction/POI guides, \texttt{P1, P2,\dots} (blue) = candidate pins / parking / depots, and \texttt{D1, D2,\dots} (slate) = demand points. Task-specific overlays extend the grammar with: icon glyphs for typed POIs (EV charger, restroom, caf\'e, pharmacy, bank, water) and building entrances (loading dock, main, service, accessible); teal directional arrows on one-way streets; a red line with an ``$\times$'' along a closed segment; a brown line with a staircase badge on impassable stairs; price/accessibility chips (\$$N$, wheelchair symbol) under parking pins; and named striped \emph{zones} drawn in deliberately non-map colors (magenta ``EVENT ZONE'', hatched ``HAZARD ZONE'', indigo ``BUILDING'') so they are not confused with basemap parks or water. A per-task legend strip names every symbol used in the panel.

\subsection{Visual-consistency invariants}
Because the model must read labels precisely, generation enforces strict invariants at \emph{every} zoom before an instance is accepted; failing any one triggers regeneration. (i) \emph{All-markers-visible}: every task-critical marker projects inside the panel bounds; we place all such markers within a $130$\,m-radius window of the center so they fit the tighter local panel with label margin, and a per-panel visibility test records exactly which IDs project on-frame so the consistency protocol skips any zoom that lacks the evidence to answer. (ii) \emph{Dot-collision-free}: marker dots (sized adaptively per zoom: radius $11$\,px on the mid panel, $14$\,px on the local panel) must be separated by at least the sum of their outer rings plus a $3$\,px buffer, and a global minimum center-to-center separation of $30$\,px is enforced, checked independently at each zoom so coarse panels are never visually mushed together. (iii) \emph{Label-collision-free}: each printed ID is placed by a layout solver that tries six candidate offsets around its dot (upper/lower-right, upper/lower-left, straight up/down), treating other dots, already-placed labels, and the reserved bounding boxes of attribute/name chips and stairs badges as obstacles; labels are drawn as bold colored text on a white halo so they remain legible over any basemap. (iv) \emph{Cue-on-top}: closure, stairs, one-way, and zone cues are composited \emph{after} the markers so the central constraint is never occluded. Together these invariants guarantee that a careful reader (human or model) can recover every label and cue at both scales, isolating reasoning failures from rendering artifacts.

\subsection{Hidden graph and exact oracles}
Each task is scored against an exact oracle computed on $G$. Shortest and constrained-shortest paths use Dijkstra's algorithm~\cite{dijkstra1959} with a forbidden-edge/node weight callable (no graph copies); near-optimality margins use Yen's $k$-shortest paths~\cite{yen1971kshortest} on a simple-digraph projection of the multigraph. Multi-stop sequencing (e.g.\ POI routing through a waypoint class) uses Held--Karp dynamic programming~\cite{heldkarp1962}. Facility location uses an exact $1$-median~\cite{hakimi1964pmedian} (and a minimax variant); service-area siting uses single-facility max-coverage~\cite{churchrevelle1974maxcover}; reachability uses single-source shortest-path distances with a margin-aware radius. Oracle outputs include the optimal cost, the second-best cost (for tie/margin checks), and the projected \emph{visible route} (the ordered marker sequence a correct answer should produce). None of $G$, $\sigma$, or $y^\star$ is exposed in the PVD track.

\subsection{Deterministic, anti-gaming generation}
The entire dataset is regenerable from a single epoch seed and the per-metro bounding boxes; all randomness uses a SHA-256-derived stable seed (not Python's randomized hash), so generation is reproducible across machines and processes. Crucially, generation enforces per-task \emph{anti-gaming} rejection filters so that the visual shortcut is never the answer. A legal route must require its waypoint detour: the start-to-waypoint-to-goal path must be at least $5\%$ longer than the direct start-to-goal path, so the trivial endpoint pair is both illegal (it skips the required waypoint) and suboptimal, and the purple guide markers deliberately mix on-route junctions with off-route distractors. A pin-placement instance is rejected whenever the Euclidean-nearest candidate coincides with the network $1$-median, forcing genuine graph-distance reasoning rather than image proximity. One-way instances require the directed-legal route to be at least $5\%$ longer than the route that would be allowed if arrows were ignored, with a mandatory waypoint reachable only by obeying the arrows. Closure and symbol-constrained instances are rejected unless the drawn constraint forces a real detour, and a scorer-consistency guard discards any instance whose own oracle route would cross the drawn constraint. To guarantee a unique, checkable answer, selection tasks enforce an explicit margin between the best and second-best option: $5\%$ for pin placement, a clean integer distance gap inside a $12\%$ band for reachability, and a $12\%$ relative gap for POI-goal routing, loading-dock parking, and the disambiguation candidates; the latter additionally require the two same-named POIs to be separated by at least $70$\,m along the query axis so the spatial relation is unambiguous. Finally, the binding of a typed-POI marker to its category is stored only in the hidden snap function, never drawn as text, so a graph- or text-only baseline cannot recover icon semantics and must read the rendered glyph.

\subsection{Real vs.\ synthesized content}
The spatial substrate is always real OSM: street graphs, POI locations and categories, building footprints, stairs (\texttt{highway=steps}), and one-way directions are taken directly from OpenStreetMap. A small set of \emph{attributes} that OSM does not record at scale are deterministically synthesized and flagged in the hidden record: parking price and wheelchair-accessibility for attribute-aware parking, entrance \emph{types} for loading-dock selection, and the drawn hazard/event \emph{zones} for symbol-constrained routing and visual disambiguation. Synthesis is seeded and never changes the underlying geometry, so the visual and graph substrate remains faithful while the attribute-fusion tasks remain exactly scorable.

\subsection{Coverage and provenance}
The benchmark spans ten U.S.\ downtowns (Seattle, New York, San Francisco, Chicago, Los Angeles, Washington DC, Boston, Portland, Denver, Houston), 12 tasks, and 6{,}000 instances ($\approx$500 per task) rendered at two zooms for 12{,}000 panels. Three tasks depend on scarce real POI configurations and therefore use metro subsets where those configurations exist: POI-goal routing (8 metros), symbol-constrained routing (7), and visual disambiguation (4, the downtowns with multiple same-named chain POIs straddling a landmark). Every panel records its style and graph version plus a style hash for provenance and exact reproduction.

\section{Tasks}
\label{sec:tasks}
\begin{figure*}[t]
\centering
\setlength{\tabcolsep}{2pt}
\renewcommand{\arraystretch}{0.7}
\begin{tabular}{cccc}
\includegraphics[width=0.235\textwidth]{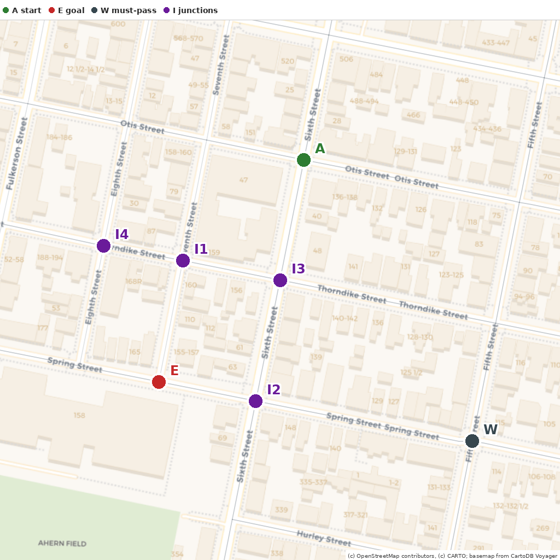} &
\includegraphics[width=0.235\textwidth]{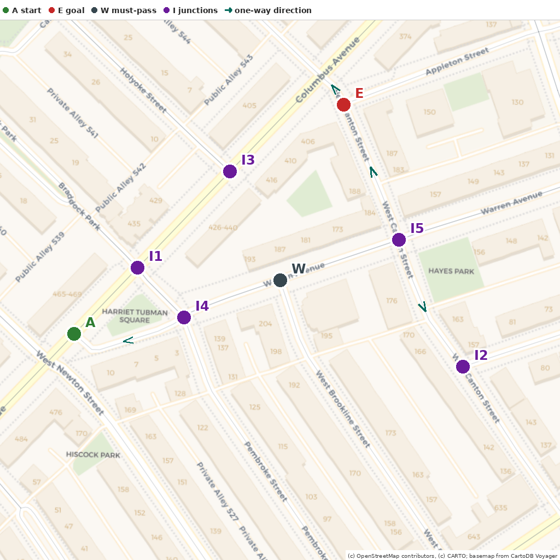} &
\includegraphics[width=0.235\textwidth]{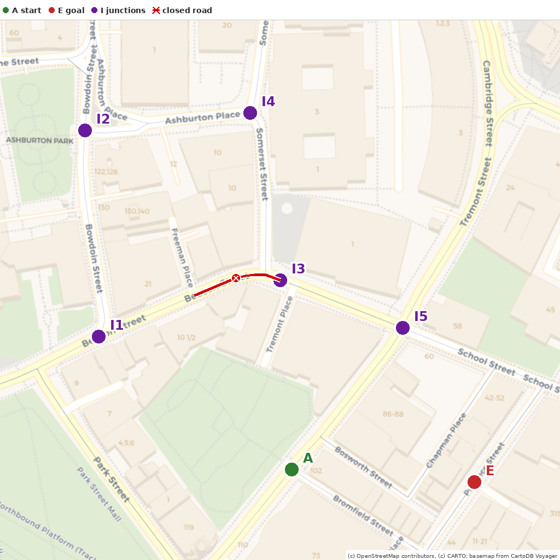} &
\includegraphics[width=0.235\textwidth]{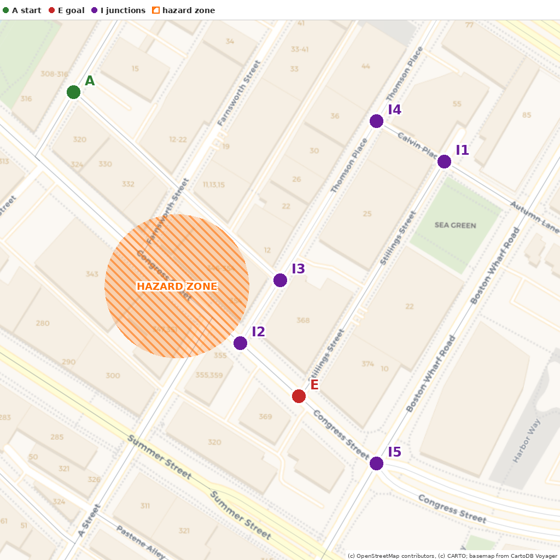} \\
{\scriptsize Legal route} & {\scriptsize One-way compliance} & {\scriptsize Closure replanning} & {\scriptsize Symbol-constrained} \\
\includegraphics[width=0.235\textwidth]{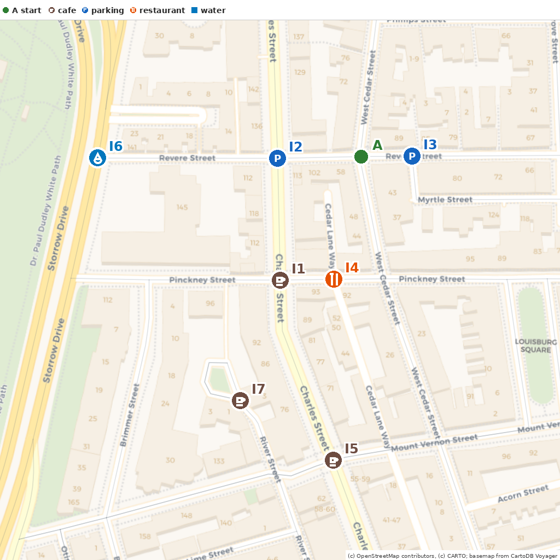} &
\includegraphics[width=0.235\textwidth]{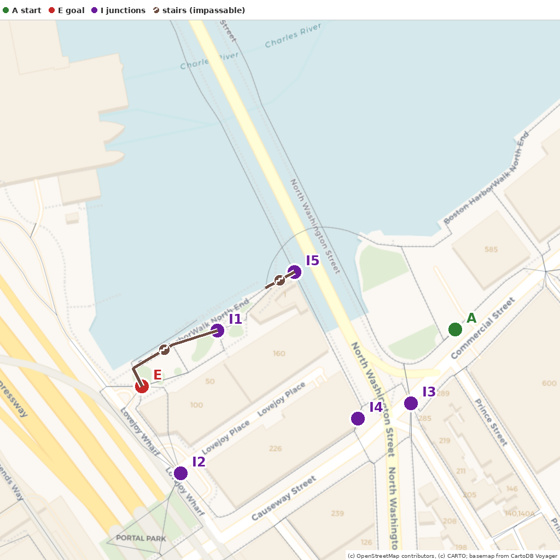} &
\includegraphics[width=0.235\textwidth]{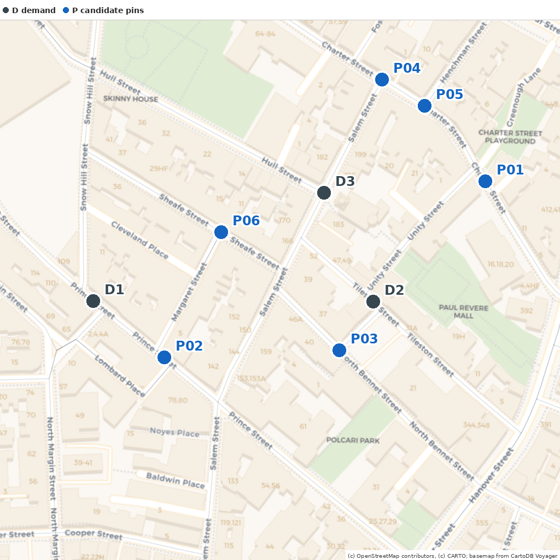} &
\includegraphics[width=0.235\textwidth]{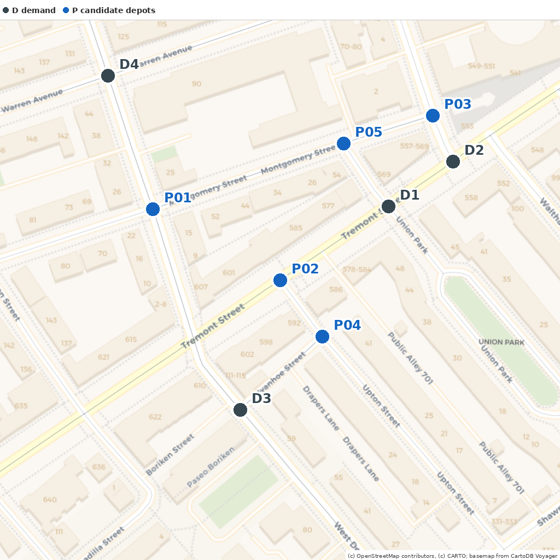} \\
{\scriptsize POI-goal routing} & {\scriptsize Step-free accessibility} & {\scriptsize Pin placement} & {\scriptsize Service-area siting} \\
\includegraphics[width=0.235\textwidth]{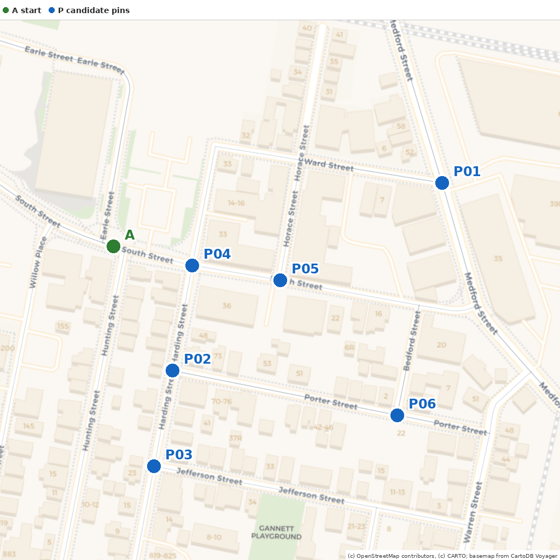} &
\includegraphics[width=0.235\textwidth]{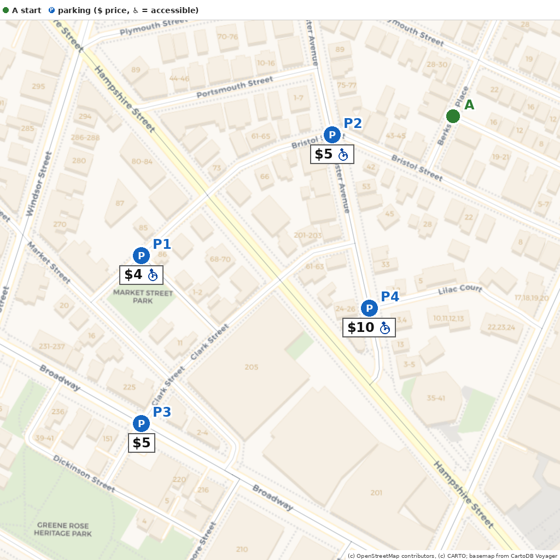} &
\includegraphics[width=0.235\textwidth]{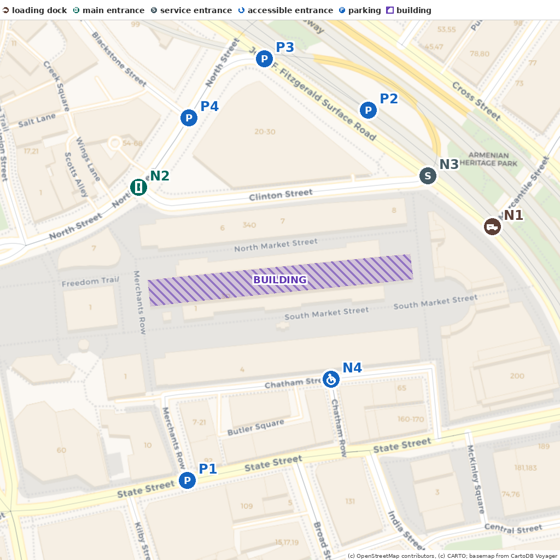} &
\includegraphics[width=0.235\textwidth]{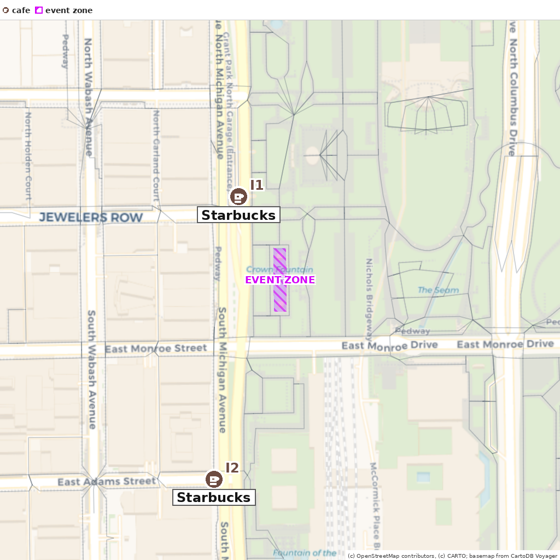} \\
{\scriptsize Reachability} & {\scriptsize Attribute-aware parking} & {\scriptsize Loading-dock selection} & {\scriptsize Visual disambiguation} \\
\end{tabular}
\caption{Representative local-zoom panels for all 12 tasks (one instance each). Every panel is self-rendered from OSM with the fixed marker grammar, task-specific symbol overlays (one-way arrows, closure $\times$, stairs badges, price/accessibility chips, named striped zones, typed POI icons), and a per-task legend. The hidden street graph and oracle answer are never shown to the model.}
\label{fig:gallery}
\end{figure*}

The 12 tasks span three families; Table~\ref{tab:tasks} summarizes their network, output, oracle, and primary metric, and Figure~\ref{fig:gallery} shows a representative panel for each. All routes are reported as an ordered list of visible marker IDs, which we snap back to the graph for scoring.

\subsection{Routing tasks}
\textbf{Legal route.} Shortest legal route from \texttt{A} to \texttt{E} that passes a required waypoint \texttt{W}, among purple junction guides that mix on-route nodes and distractors. The waypoint-detour filter guarantees the naive $[\texttt{A},\texttt{E}]$ answer is both illegal and suboptimal.
\textbf{One-way compliance.} Shortest legal \emph{driving} route on the directed graph that obeys every drawn one-way arrow and passes a waypoint reachable only by respecting arrow directions; the directed optimum is forced to be meaningfully longer than the wrong-way-allowed route.
\textbf{Closure replanning.} Shortest route avoiding a road segment drawn as a red ``$\times$''; the model must also report whether \texttt{E} remains reachable.
\textbf{Step-free accessibility.} Shortest route for a wheelchair/walker profile that avoids all drawn stairs (real \texttt{highway=steps} edges); stair-dense windows are rejected to keep panels legible.
\textbf{Symbol-constrained routing.} Shortest route from \texttt{A} to \texttt{E} avoiding a drawn closure \emph{and} a striped hazard zone; the constraints exist only as rendered symbols, never as graph attributes, so the model must \emph{see} them.
\textbf{POI-goal routing.} From \texttt{A}, route to the nearest goal-type icon (e.g.\ EV charger) while passing at least one waypoint-type icon (e.g.\ restroom); the model must first identify the correct icons visually, then route, so a graph-only baseline cannot pick the node.

\subsection{Facility and set-selection tasks}
\textbf{Pin placement.} Choose the candidate pin minimizing total network distance to the demand markers ($1$-median~\cite{hakimi1964pmedian}); the Euclidean-nearest pin is filtered out as the answer.
\textbf{Service-area siting.} Choose the depot covering the most demand markers within a network-distance budget (single-facility max-coverage~\cite{churchrevelle1974maxcover}).
\textbf{Reachability.} List every candidate pin reachable from \texttt{A} within a stated network-distance budget; the radius is snapped to an integer inside a clean distance gap so the set is unambiguous.
\textbf{Attribute-aware parking.} Choose the cheapest wheelchair-accessible parking within a budget of \texttt{A}, fusing the price chip, the accessibility symbol, and the graph distance.
\textbf{Loading-dock selection.} For an indigo striped \emph{building} with several entrance icons, pick the loading-dock entrance and the nearest parking lot by street distance; entrance types are synthesized while building and parking are real.

\subsection{Visual disambiguation}
Two POIs share the same name (icons with name chips). Given a named striped zone, the model selects the POI in the queried spatial relation to it (e.g.\ the one east of, or across the river from, the zone). This isolates visual-spatial reasoning: a name lookup or a graph-cost computation alone cannot solve it.

\begin{table}[t]
\centering
\caption{The 12 tasks. Net = hidden graph (D: directed drive, W: walk). Output: route (ordered marker IDs), pick (single ID), set (ID set), pair (two IDs). Primary metric reported in Table~\ref{tab:leaderboard}.}
\label{tab:tasks}
\footnotesize
\begin{tabular}{l c l l}
\toprule
Task & Net & Output & Primary metric \\
\midrule
Legal route            & D & route & legal-route rate \\
One-way compliance     & D & route & legal-route rate \\
Closure replanning     & D & route & legal-route rate \\
Symbol-constrained     & D & route & legal-route rate \\
POI-goal routing       & D & route & legal-route rate \\
Step-free accessibility& W & route & legal-route rate \\
Pin placement          & D & pick  & exact match \\
Service-area siting     & D & pick  & exact match \\
Attribute-aware parking & D & pick  & exact match \\
Reachability           & D & set   & set-F1 \\
Loading-dock selection & D & pair  & exact match \\
Visual disambiguation  & W & pick  & exact match \\
\bottomrule
\end{tabular}
\end{table}

\section{Evaluation Protocol}
\label{sec:protocol}
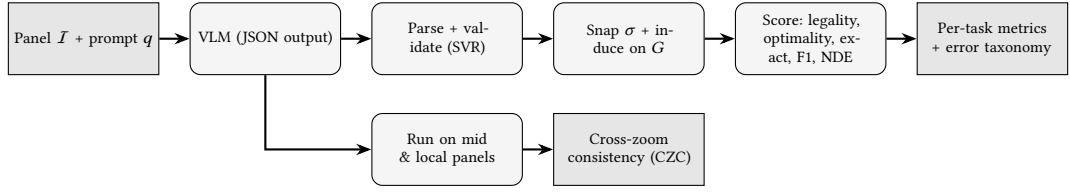
\begin{figure*}[t]
\centering
\begin{tikzpicture}[node distance=0.42cm and 0.40cm]
  \node[io] (e1) {Panel $\mathcal{I}$ + prompt $q$};
  \node[stage, right=of e1] (e2) {VLM (JSON output)};
  \node[stage, right=of e2] (e3) {Parse + validate (SVR)};
  \node[stage, right=of e3] (e4) {Snap $\sigma$ + induce on $G$};
  \node[stage, right=of e4] (e5) {Score: legality, optimality, exact, F1, NDE};
  \node[io, right=of e5] (e6) {Per-task metrics + error taxonomy};
  \node[stage, below=0.5cm of e3] (c1) {Run on mid \& local panels};
  \node[io, right=of c1] (c2) {Cross-zoom consistency (CZC)};
  \draw[flow] (e1)--(e2); \draw[flow] (e2)--(e3); \draw[flow] (e3)--(e4);
  \draw[flow] (e4)--(e5); \draw[flow] (e5)--(e6);
  \draw[flow] (e2.south) |- (c1); \draw[flow] (c1)--(c2);
\end{tikzpicture}
\caption{Evaluation workflow (Pure Visual Decision track). The model sees only $(\mathcal{I},q)$; its JSON answer passes a strict gating chain (schema validity, then snapping and induction on the hidden graph, then task scoring). The lower branch runs the model independently on both zoom panels to measure cross-zoom consistency (CZC).}
\label{fig:eval}
\end{figure*}

We evaluate a \emph{Pure Visual Decision} (PVD) track: the model sees only the rendered panel(s) and a prompt containing the marker grammar and the exact JSON answer schema, and never sees the graph, the snapping function, or a manifest. Scoring proceeds as a strict chain so that each stage gates the next; Figure~\ref{fig:eval} summarizes this evaluation workflow.

\subsection{Prompting and structured decoding}
Every model is queried with a fixed two-part prompt (Figure~\ref{fig:prompt}). A \emph{system} prompt teaches the visual grammar (marker colors and IDs, the faint street-graph overlay, the legend, and the special cues for closures, stairs, one-way arrows, and named zones) and the global rules: travel only along drawn streets; ``distance'' always means street-network distance, never straight-line; use marker IDs exactly as printed; a route asked from \texttt{A} to \texttt{E} must begin at \texttt{A} and end at \texttt{E}; reason internally but output \emph{only} a single JSON object; and abstain honestly when a needed label is illegible or absent rather than guess. The \emph{user} prompt then adds a one-line zoom note, the instance question, task-specific solving guidance, and a worked example of the exact JSON to return. Answers use an envelope \texttt{\{task, answer, abstain, confidence\}}; the abstain flag and confidence let a model decline rather than fabricate, and are recorded separately from incorrect answers. We request structured output in each provider's strongest available mode (a per-task JSON \emph{schema} where supported, otherwise a generic JSON-object mode), and validate every response with a per-task Pydantic model that ignores extraneous keys and also accepts a bare answer object; before declaring a response invalid, a tolerant parser first recovers well-formed JSON from fenced or prose-wrapped completions. Because the prompt never reveals the snap function $\sigma$ (the mapping from a visible ID to its graph node, and for typed POIs its hidden category), the decision cannot be produced without actually reading the panel.

\begin{figure}[t]
\footnotesize
\setlength{\fboxsep}{6pt}
\noindent\fbox{\begin{minipage}{0.92\columnwidth}
\textbf{System (excerpt).} You are an expert cartographic reasoning agent. You read a single rendered street-map panel and output an exact, machine-checkable mobility decision. Markers are colored dots with a bold printed ID: \texttt{A}=green start, \texttt{E}=red goal, \texttt{W}=slate waypoint, \texttt{I\#}=purple junction/POI guides, \texttt{P\#\#}=blue candidate pins/depots, \texttt{D\#}=slate demand points. A legend strip names every symbol used in the panel. Rules: travel only along drawn streets; ``distance'' means street-network distance; use IDs exactly as printed; output only a single JSON object; abstain honestly if a needed label is unreadable.

\smallskip
\textbf{User (pin placement).} \emph{(local panel; labels are large and legible.)} TASK: Pick the one pin (\texttt{P\#\#}) that minimizes total street-network distance to demand markers \texttt{D1}, \texttt{D2}, \texttt{D3}. HOW TO SOLVE: this is a 1-median choice; the straight-line-nearest pin is often not the network-optimal one, so reason about the roads. Output ONLY this JSON shape:

\smallskip
{\ttfamily \{"task":"pin\_placement", "answer":\{"selected\_pin\_id":"P04"\}, "abstain":false, "confidence":0.7\}}
\end{minipage}}
\caption{Example Pure Visual Decision prompt (pin placement). The system prompt fixes the visual grammar and global rules; the user prompt adds the zoom note, the instance task, task-specific guidance, and the exact JSON answer shape. The hidden graph, snap function $\sigma$, and oracle are never shown to the model.}
\label{fig:prompt}
\end{figure}

\textbf{(1) Schema validity (SVR).} The model returns a JSON envelope; we parse it (tolerating fenced/prose-wrapped JSON) and validate against the task's schema. Schema validity is the first scored event: malformed output zeros all downstream metrics. Parsing is crash-safe (a malformed envelope is recorded as schema-invalid, not an error).

\textbf{(2) Snapping and induction.} Valid answers reference visible IDs; we apply $\sigma$ to map each ID to its graph node and \emph{induce} the decision on $G$ (e.g.\ expand a marker sequence into the concrete edge path between consecutive nodes). Route answers must begin at \texttt{A} and end at \texttt{E}; otherwise they are flagged \emph{incomplete} and earn no legality/optimality credit.

\textbf{(3) Task scoring.} For routes we report: \emph{legality} (every induced edge exists and obeys the task constraints: one-way directions, closures, stairs, hazard zones, and required waypoints), the \emph{optimality ratio} $c_{\text{pred}}/c^\star$ and \emph{regret} $c_{\text{pred}}-c^\star$ versus the oracle cost (finite only for legal routes), \emph{edge-IoU} against the oracle edge set, and \emph{normalized edit distance} over the marker sequence. For single selections we report \emph{exact match} and, for pins/docks, the \emph{network-distance error} between the chosen and optimal node. For sets (reachability) we report \emph{precision}, \emph{recall}, and \emph{F1}; for siting, the \emph{coverage ratio} and \emph{coverage regret}. An error taxonomy attributes failures to schema, symbol-grounding (a chosen ID is not a valid candidate), incomplete route, illegality, or sub-optimality.

\textbf{(4) Cross-zoom consistency (CZC).} We run the model independently on the mid and local panels of the same instance and measure decision agreement: edge-Jaccard for routes, set-Jaccard for reachability, and exact agreement for single picks. CZC is computed \emph{independently of correctness} (a model can be accurate yet scale-inconsistent), making it a distinct axis from accuracy.

\section{Experimental Setup}
\label{sec:setup}
\textbf{Models and APIs.} We evaluate two tiers of vision-language models behind a single OpenAI-compatible client: a \emph{fast} tier (Google Gemini Flash-Lite variants, OpenAI GPT-4o-mini, Anthropic Claude Haiku, and an open-weights Gemma served on our own infrastructure) and a \emph{frontier} reasoning tier (Anthropic Claude Opus and GPT-5.5). All providers (OpenAI, Anthropic, Google Gemini, and the self-hosted Gemma endpoint) are reached through one uniform chat-completions interface, so the same prompt builder, image encoder, and answer parser drive every model and adding a new model is a single registry entry that declares its provider, structured-output capability, and reasoning controls.

\textbf{Harness and concurrency.} The evaluation harness is fully asynchronous (Python \texttt{asyncio} over an \texttt{AsyncOpenAI} client). Requests are dispatched concurrently under a global semaphore that bounds the number of in-flight calls, while a round-robin rotation across multiple API keys spreads load and respects per-key rate limits. Each call is wrapped in bounded retry with exponential backoff and jitter: transient network errors, rate-limit responses, and empty completions are retried, and unsupported request fields (for example, a rejected \texttt{response\_format} or a reasoning-effort flag) are detected and dropped before the call is re-issued, so one model's API quirks never abort the run. Predictions are streamed to disk incrementally, making runs fully resumable and idempotent (already-graded instances are skipped on restart).

\textbf{Structured output and schema validation.} Every task declares an exact JSON answer schema. We request structured output in each provider's strongest available mode (native JSON-schema where supported, otherwise JSON-object), and validate each response against a per-task Pydantic model. Schema validity is the first gate of the scoring chain: a response that fails to parse or violates the schema is recorded as schema-invalid rather than crashing the harness, and a tolerant fallback parser first attempts to recover well-formed JSON from fenced or prose-wrapped completions. This separates output-formatting failures from genuine reasoning errors.

\textbf{Protocol.} All models see identical panels, the same marker grammar, and the same answer schema; none is given the hidden graph, the snapping function, or a manifest. Prompts include an explicit abstain option so a model can decline rather than guess, and the reasoning tier exposes a configurable reasoning-effort setting that we tune for tractable latency. Generation is seeded for reproducibility, and the full harness, prompts, schemas, and model registry are released so any model can be re-run on the same inputs.

\section{Results}
\label{sec:results}
\begin{figure}[t]
\centering
\includegraphics[width=\columnwidth]{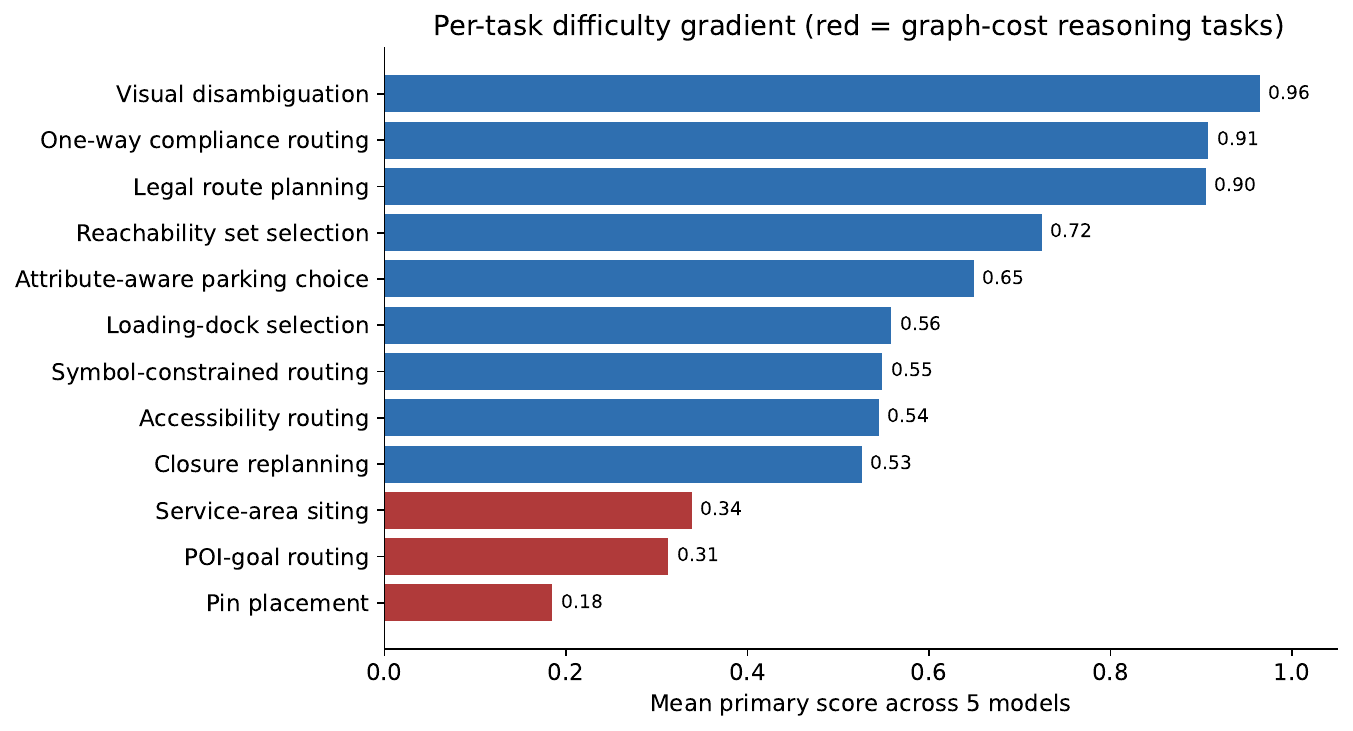}
\caption{Per-task difficulty gradient: mean primary score across the seven models. Graph-cost reasoning tasks (red) sit at the bottom; pin placement is near chance.}
\label{fig:gradient}
\end{figure}

\begin{figure}[t]
\centering
\includegraphics[width=\columnwidth]{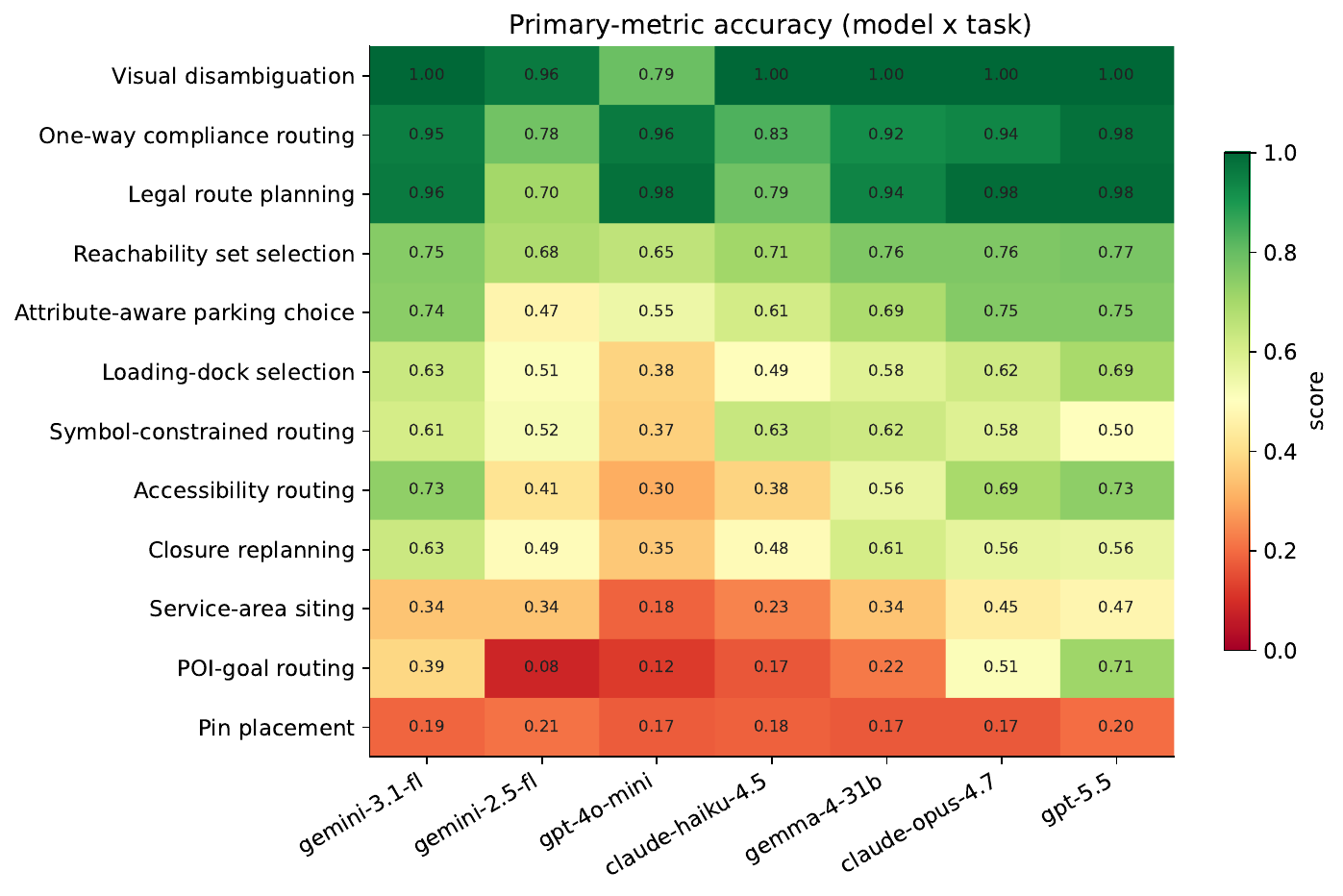}
\caption{Primary accuracy for every model and task. The bottom row (pin placement) is uniformly poor across all seven models, including the two frontier reasoning models (right).}
\label{fig:heatmap}
\end{figure}

\begin{figure}[t]
\centering
\includegraphics[width=\columnwidth]{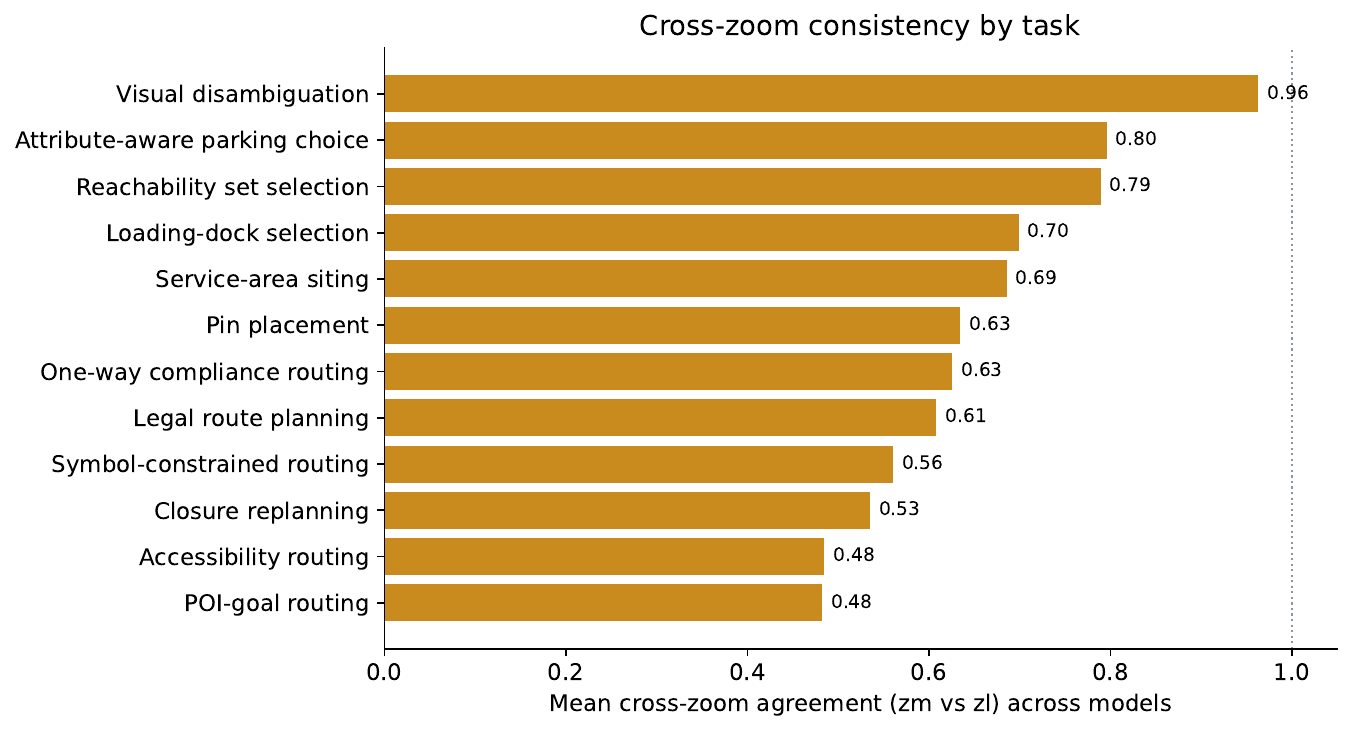}
\caption{Cross-zoom consistency by task (mean over models): agreement between the wide and local zoom decisions on the same instance. Routing decisions frequently flip across scale even when accuracy is high.}
\label{fig:czc}
\end{figure}

Table~\ref{tab:leaderboard} reports per-task primary accuracy for all seven models; Figure~\ref{fig:gradient} shows the resulting difficulty gradient, Figure~\ref{fig:heatmap} the full model$\times$task grid, and Figure~\ref{fig:czc} cross-zoom consistency by task.

\textbf{VLMs read maps and route simply.} Visual disambiguation is near-solved (0.79--1.00) and legal/one-way routing is strong (0.70--0.98): models reliably read the marker grammar and trace short, connected routes.

\textbf{They collapse on graph-cost reasoning.} The difficulty gradient falls off sharply for tasks that require comparing distances \emph{along} the network. Most strikingly, single-facility pin placement (1-median) is near chance for \emph{every} model (0.17--0.21, including both frontier reasoning models), because the visually nearest candidate is, by construction, often not the network-optimal one. Service-area siting (0.19--0.47) and POI-goal routing (0.08--0.71) are similarly weak in the fast tier.

\textbf{Frontier reasoning helps some tasks but not the core failure.} GPT-5.5 lifts POI-goal routing from 0.39 (best fast) to 0.71 and service-area siting from 0.35 to 0.47, and leads the overall macro accuracy (0.70). Yet pin placement remains at chance, 0.20 for GPT-5.5 and 0.17 for Claude Opus; this indicates the bottleneck is graph-cost optimization, not perception or model scale.

\textbf{Decisions are scale-inconsistent.} Even the best model agrees with itself across zoom only $\approx$73\% of the time (macro CZC, Table~\ref{tab:leaderboard}); for routing tasks CZC is often 0.52--0.67 despite high accuracy. High accuracy does not imply scale-invariance, motivating CZC as a distinct axis.

\begin{table*}[t]
\centering
\caption{Per-task primary score (0--1) for all seven models (mean over both zooms). Primary metric: legal-route rate (routing tasks), set-F1 (reachability), exact selection (pin/service/parking/dock/visual). Tasks ordered by mean difficulty. \textbf{Bold} = best per row; ($\dagger$) marks the frontier reasoning tier. The bottom rows give macro accuracy and macro cross-zoom consistency (CZC).}
\label{tab:leaderboard}
\footnotesize
\begin{tabular}{l ccccc cc}
\toprule
Task & Gemini-3.1-FL & Gemini-2.5-FL & GPT-4o-mini & Claude-Haiku-4.5 & Gemma-4-31B & Claude-Opus-4.7$^\dagger$ & GPT-5.5$^\dagger$ \\
\midrule
Visual disambiguation        & \textbf{1.00} & 0.96 & 0.79 & \textbf{1.00} & \textbf{1.00} & \textbf{1.00} & \textbf{1.00} \\
One-way compliance routing   & 0.95 & 0.78 & 0.96 & 0.83 & 0.92 & 0.94 & \textbf{0.98} \\
Legal route planning         & 0.96 & 0.70 & \textbf{0.98} & 0.79 & 0.94 & \textbf{0.98} & \textbf{0.98} \\
Reachability set selection   & 0.75 & 0.68 & 0.65 & 0.71 & 0.76 & 0.76 & \textbf{0.77} \\
Attribute-aware parking      & 0.74 & 0.47 & 0.55 & 0.61 & 0.69 & \textbf{0.75} & \textbf{0.75} \\
Loading-dock selection       & 0.63 & 0.51 & 0.38 & 0.49 & 0.58 & 0.62 & \textbf{0.70} \\
Symbol-constrained routing   & 0.61 & 0.52 & 0.37 & \textbf{0.63} & 0.62 & 0.59 & 0.50 \\
Step-free accessibility      & 0.73 & 0.41 & 0.30 & 0.38 & 0.56 & 0.70 & \textbf{0.74} \\
Closure replanning           & \textbf{0.63} & 0.49 & 0.35 & 0.49 & 0.61 & 0.57 & 0.56 \\
Service-area siting          & 0.35 & 0.35 & 0.19 & 0.24 & 0.35 & 0.45 & \textbf{0.47} \\
POI-goal routing             & 0.39 & 0.08 & 0.12 & 0.17 & 0.22 & 0.51 & \textbf{0.71} \\
Pin placement (1-median)     & 0.19 & \textbf{0.21} & 0.18 & 0.18 & 0.17 & 0.17 & 0.20 \\
\midrule
\textbf{Macro accuracy}      & 0.66 & 0.51 & 0.48 & 0.54 & 0.62 & 0.67 & \textbf{0.70} \\
\textbf{Macro CZC}           & \textbf{0.73} & 0.62 & 0.59 & 0.60 & 0.61 & 0.71 & 0.72 \\
\bottomrule
\end{tabular}
\end{table*}

\section{Discussion and Open Challenges}
\label{sec:discussion}
The results separate two abilities that prior text-scored benchmarks conflate. VLMs are competent map \emph{readers}: they parse a fixed marker grammar, recognize icons, and resolve spatial relations (visual disambiguation near 1.0). They are poor map \emph{reasoners} whenever the answer depends on comparing distances along the network. Pin placement is the cleanest probe: it is a single discrete choice with no long-horizon planning, yet accuracy is at chance for all seven models because the model must integrate road-network distance rather than image proximity. This points to a concrete, industrially relevant open challenge (grounding facility-location and least-cost reasoning in the visible graph) that better perception alone will not solve. Cross-zoom inconsistency compounds the problem for deployment: a system that changes its route when the map is zoomed is hard to trust. We release the benchmark and harness at \url{https://github.com/Vi-Sri/mapreason-osm} to make progress on these axes measurable.

\section{Limitations}
Three tasks use metro subsets due to scarce real POI configurations, so per-task metro coverage is non-uniform. We evaluate a single visual-only prompting track; manifest-assisted and candidate-rerank tracks (to separate perception, grounding, and planning failures) are future work. Finally, model identifiers and endpoints evolve; we release the deterministic generator and harness so results can be reproduced and extended.

\section{Conclusion}
\label{sec:conclusion}
MapReason-OSM is a graph-grounded, exactly-scored benchmark that asks whether VLMs can turn a street-map image into a legal, near-optimal mobility decision. Across seven models, the answer is nuanced: models read maps and route simply, but fail at graph-cost reasoning (pin placement is at chance even for frontier reasoning models), and are frequently inconsistent across zoom. By isolating where map perception ends and spatial reasoning fails, and by shipping a reproducible generator and harness, MapReason-OSM offers the community a precise target for spatially grounded foundation models.

\begin{acks}
Generative AI tools were used to assist with code, figure generation, and manuscript drafting; all content was reviewed and verified by the authors. Map data is \textcopyright{} OpenStreetMap contributors, available under the Open Database License (ODbL).
\end{acks}

\bibliographystyle{ACM-Reference-Format}
\bibliography{references}

\end{document}